\documentclass[10pt,twocolumn,letterpaper]{article}

\usepackage{chngcntr} 

\usepackage{iccv}
\usepackage{times}
\usepackage{epsfig}
\usepackage{graphicx}
\usepackage{amsmath}
\usepackage{amssymb}
\usepackage{graphicx}
 \usepackage{multirow}
\usepackage{amsmath,amssymb} 
\usepackage{arydshln}

\usepackage{color}
\usepackage{cite}
\usepackage{dsfont}
\usepackage[table,xcdraw]{xcolor}
\usepackage{ragged2e}
\usepackage{nccmath}
\usepackage{textcomp}

\usepackage{csquotes}

\DeclareMathOperator*{\argmin}{arg\,min}

\usepackage[pagebackref=true,breaklinks=true,letterpaper=true,colorlinks,bookmarks=false]{hyperref}

\iccvfinalcopy 

\def\iccvPaperID{105} 

\ificcvfinal\fi
\begin{document}

\title{What Else Can Fool Deep Learning? \\ Addressing Color Constancy Errors on Deep Neural Network Performance}


\author{Mahmoud Afifi$^{1}$\\
$^{1}$York University, Toronto\\
{\tt\small mafifi@eecs.yorku.ca}
\and
Michael S Brown$^{1,2}$\\
$^{2}$Samsung AI Center, Toronto\\
{\tt\small mbrown@eecs.yorku.ca}
}

\maketitle
\ificcvfinal\thispagestyle{empty}\fi

\begin{abstract}
There is active research targeting local image manipulations that can fool deep neural networks (DNNs) into producing incorrect results.  This paper examines a type of {\it global} image manipulation that can produce similar adverse effects.  Specifically, we explore how strong color casts caused by incorrectly applied computational color constancy -- referred to as white balance (WB) in photography -- negatively impact the performance of DNNs targeting image segmentation and classification.  In addition, we discuss how existing image augmentation methods used to improve the robustness of DNNs are not well suited for modeling WB errors. To address this problem, a novel augmentation method is proposed that can emulate accurate color constancy degradation. We also explore pre-processing training and testing images with a recent WB correction algorithm to reduce the effects of incorrectly white-balanced images. We examine both augmentation and pre-processing strategies on different datasets and demonstrate notable improvements on the CIFAR-10, CIFAR-100, and ADE20K datasets.
\end{abstract}


\section{Introduction} \label{sec:introduction}

\begin{figure}
\begin{center}
\includegraphics[width=\linewidth]{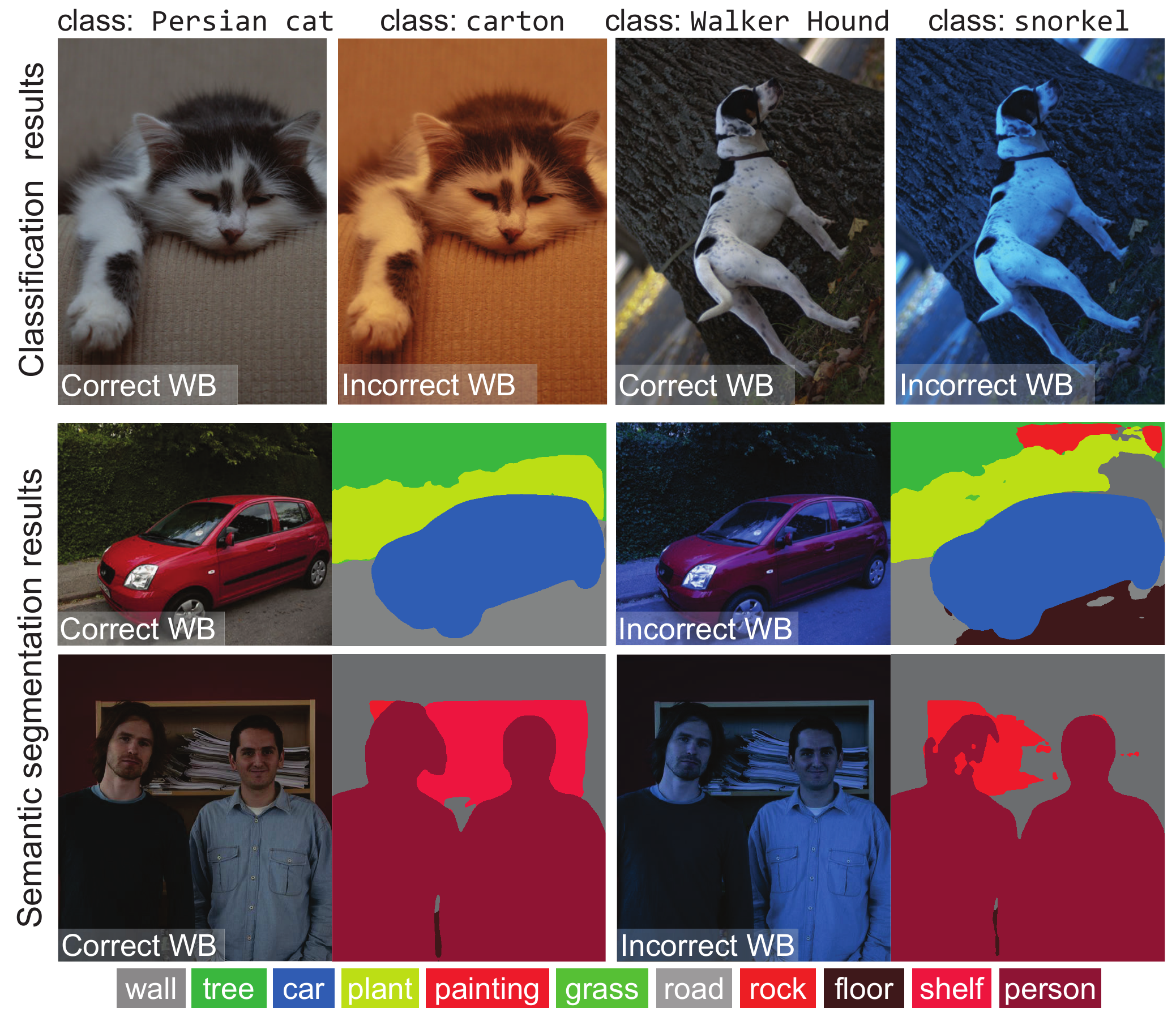}
\end{center}
\vspace{-2mm}
   \caption{The effect of correct/incorrect computational color constancy (i.e., white balance) on (top) classification results by ResNet \cite{he2016deep}; and (bottom) semantic segmentation by RefineNet \cite{lin2017refinenet}.}
\label{fig:teaser}
\end{figure}

There is active interest in local image manipulations that can be used to fool deep neural networks (DNNs) into producing erroneous results.  Such ``adversarial attacks'' often result in drastic misclassifications.  We examine a less explored problem of \textit{global} image manipulations that can result in similar adverse effects on DNNs' performance.  In particular, we are interested in the role of computational color constancy, which makes up the white-balance (WB) routine on digital cameras.

We focus on computational color constancy because it represents a common source of global image errors found in real images.   When WB is applied incorrectly on a camera, it results in an undesirable color cast in the captured image. Images with such strong color casts are often discarded by users.  As a result, online image databases and repositories are biased to contain mostly correctly white-balanced images. This is an implicit assumption that is not acknowledged for datasets composed of images crawled from the web and online.  However, in real-world applications, it is unavoidable that images will, at some point, be captured with the incorrect WB applied.  Images with incorrect WB can have unpredictable results on DNNs trained on white-balanced biased training images, as demonstrated in Fig.~\ref{fig:teaser}.

\paragraph{Contribution}~We examine how errors related to computational color constancy can adversely affect DNNs focused on image classification and semantic segmentation.  In addition, we show that image augmentation strategies used to expand the variation of training images are not well suited to mimic the type of image degradation caused by color constancy errors.  To address these problems, we introduce a novel color augmentation method that can accurately emulate realistic color constancy degradation. We also examine a newly proposed WB correction method~\cite{afifi2019color} to pre-process testing and training images. Experiments on CIFAR-10, CIFAR-100, and the ADE20K datasets using the proposed augmentation and pre-processing correction demonstrate notable improvements to test image inputs with color constancy errors. Code for our proposed color augmenter is available at: \href{https://github.com/mahmoudnafifi/WB_color_augmenter}{https://github.com/mahmoudnafifi/WB\_color\_augmenter}.

\section{Related Work} \label{sec:relatedwork}

\paragraph{Computational Color Constancy}Cameras have onboard image signal processors (ISPs) that convert the raw-RGB sensor values to a standard RGB output image (denoted as an sRGB image)~\cite{ramanath2005color, karaimer2016software}. Computational color constancy, often referred to as WB in photography, is applied to mimic the human's ability to perceive objects as the same color under any type of illumination.   WB is used to identify the color temperature of the scene's illumination either manually or automatically by estimating the scene's illumination from an input image (e.g., \cite{buchsbaum1980spatial, SoG, barron2015convolutional,cheng2015effective,shi2016deep, barron2017fast,hu2017fc4, afifi2019sensor}). After WB is applied to the raw-RGB image, a number of additional nonlinear photo-finishing color manipulations are further applied by the ISP to {\it render} the final sRGB image \cite{afifi2019color}.  These photo-finishing operations include, but are not limited to, hue/saturation manipulation, general color manipulation, and local/global tone mapping~\cite{ramanath2005color, karaimer2016software, hasinoff2016burst, nam2017modelling, brooks2018unprocessing}. Cameras generally have multiple photo-finishing styles the user can select~\cite{kim2012new, karaimer2016software, afifi2019color}.

\paragraph{Post-WB Correction in sRGB Images} When WB is applied incorrectly, it results in sRGB images with strong color casts. Because of the nonlinear photo-finishing operations applied by the ISP after WB, correcting mistakes in the sRGB image is non-trivial~\cite{nguyen2018raw, afifi2019color}. Current solutions require meta-data, estimated from radiometric calibration or raw-image reconstruction methods~(e.g., \cite{kim2012new, chakrabarti2014modeling, nguyen2018raw}),  that contains the necessary information to undo the particular nonlinear photo-finishing processes applied by the ISP.  By converting back to a raw-RGB space, the correct WB can be applied using a diagonal correction matrix and then re-rendered by the ISP.  Unfortunately, meta-data to inverse the camera pipeline and re-render the image is rarely available, especially for sRGB images gleaned from the web---as is the case with existing computer vision datasets. Recently, it was shown that white balancing sRGB images can be achieved by estimating a high-degree polynomial correction matrix \cite{afifi2019color}. The work in \cite{afifi2019color}, referred to WB for sRGB images (WB-sRGB), introduces a data-driven framework to estimate such polynomial matrix for a given testing image. We build on the WB-sRGB \cite{afifi2019color} by extending this framework to emulate WB errors on the final sRGB images, instead of correcting WB. We also used the WB-sRGB method \cite{afifi2019color} to examine applying a pre-process WB correction on training and testing images in order to improve the performance of DNN models against incorrectly white-balanced images.

\begin{figure}
\begin{center}
\includegraphics[width=0.97\linewidth]{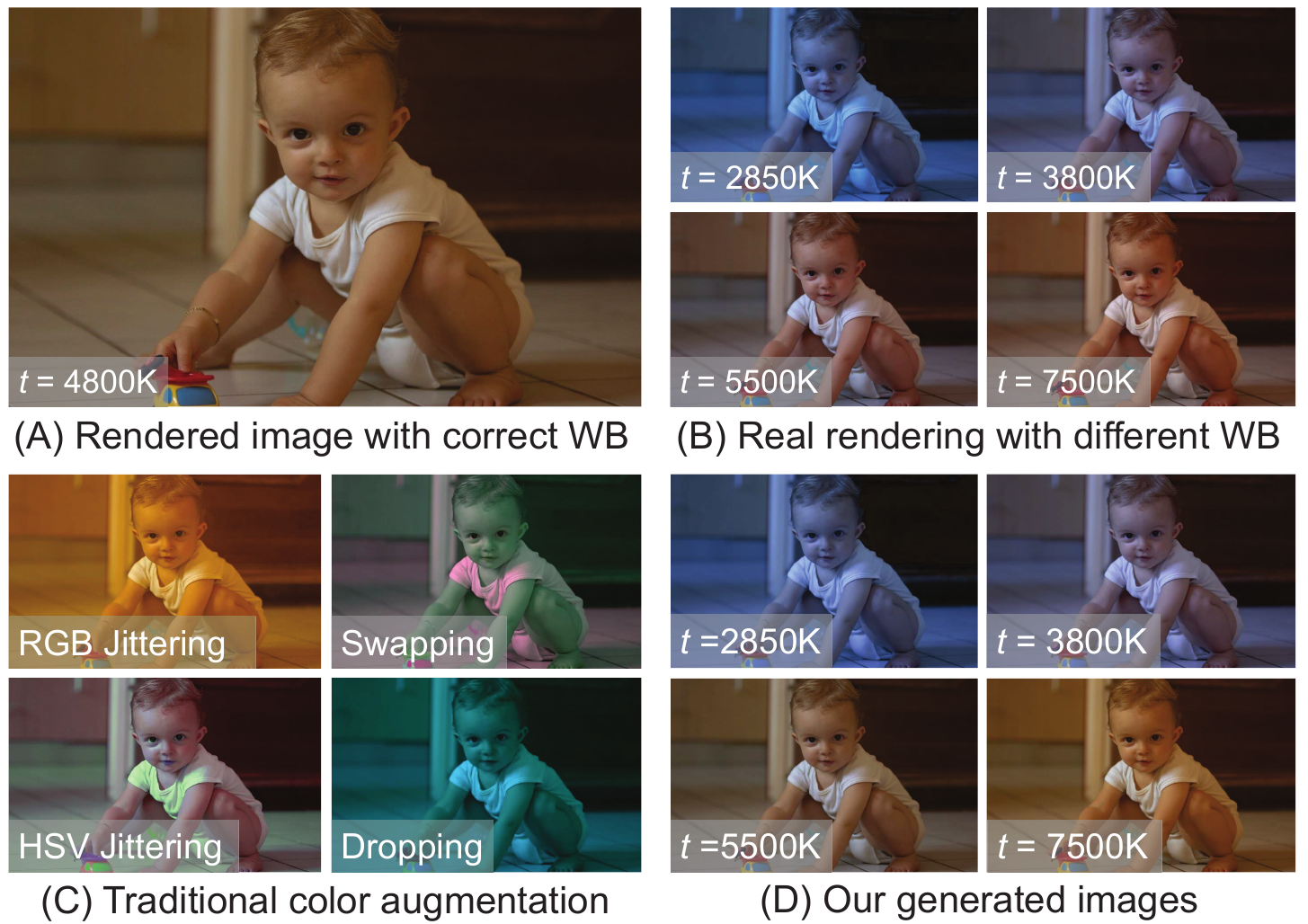}
\vspace{-1mm}
\end{center}
   \caption{(A) An sRGB image from a camera with the correct WB applied.  (B) Images from the same camera with the incorrect WB color temperatures ($t$) applied. (C) Images generated by processing image (A) using existing augmentation methods---the images clearly do not represent those in (B).  (D) Images generated from (A) using our proposed method detailed in Sec.~\ref{sec:method}.}
\label{fig:AugExamples}
\end{figure}

\paragraph{Adversarial Attacks}~DNN models are susceptible to adversarial attacks in the form of \textit{local} image manipulation (e.g., see~\cite{szegedy2013intriguing, goodfellow2014explaining, kurakin2016adversarial, cisse2017parseval}). These images are created by adding a carefully crafted imperceptible perturbation layer to the original image \cite{szegedy2013intriguing, goodfellow2014explaining}. Such perturbation layers are usually represented by \textit{local} non-random adversarial noise \cite{szegedy2013intriguing,goodfellow2014explaining,moosavi2016deepfool,xie2017adversarial, akhtar2018threat} or \textit{local} spatial transformations \cite{xiao2018spatially}. Adversarial examples are able to misguide pre-trained models to predict either a certain wrong response (i.e., targeted attack) or any wrong response (i.e., untargeted attack)~\cite{liu2017delving, 7958570,akhtar2018threat}.  While incorrect color constancy is not an explicit attempt at an adversarial attack, the types of failures produced by this \textit{global} modification act much like an untargeted attack and can adversely affect DNNs' performance.

\paragraph{Data Augmentation} To overcome limited training data and to increase the visual variation, image augmentation techniques are applied to training images. Existing image augmentation techniques include: geometric transformations (e.g., rotation, translation, shearing) \cite{hauberg2016dreaming, perez2017effectiveness, hauberg2016dreaming, cubuk2018autoaugment}, synthetic occlusions \cite{zhong2017random}, pixel intensity processing (e.g., equalization, contrast adjustment, brightness, noise) \cite{veeravasarapu2017adversarially, cubuk2018autoaugment}, and color processing (e.g., RGB color jittering and PCA-based shifting, HSV jittering,  color channel dropping, color channel swapping)~\cite{cubuk2018autoaugment,chatfield2014return, ImgaugLib, krizhevsky2012imagenet, redmon2016you, doersch2015unsupervised, movshovitz2016useful, lee2017unsupervised, kalantari2017deep}. Traditional color augmentation techniques randomly change the original colors of training images aiming for better generalization and robustness of the trained model in the inference phase. However, existing color augmentation methods often generate unrealistic colors which rarely happen in reality (e.g., green skin or purple grass). More importantly, the visual appearance of existing color augmentation techniques does not well represent the color casts produced by incorrect WB applied onboard cameras, as shown in Fig.~\ref{fig:AugExamples}.  As demonstrated in \cite{andreopoulos2012sensor, diamond2017dirty, carlson2018modeling}, image formation has an important effect on the accuracy of different computer vision tasks. Recently, a simplified version of the camera imaging pipeline was used for data augmentation \cite{carlson2018modeling}. This augmentation method in \cite{carlson2018modeling}, however, explicitly did not consider the effects of incorrect WB due to the subsequent nonlinear operations applied after WB.
To address this issue, we propose a camera-based augmentation technique that can synthetically generates images with realistic WB settings.

\paragraph{DNN Normalization Layers}
Normalization layers are commonly used to improve the efficiency of the training process. Such layers apply simple statistics-based shifting and scaling operations to the activations of network layers. The shift and scale factors can be computed either from the entire mini-batch (i.e., batch normalization \cite{ioffe2015batch}) or from each training instance (i.e., instance normalization \cite{ulyanov2017improved}). Recently, batch-instance normalization (BIN) \cite{nam2018batch} was introduced to ameliorate problems related to styles/textures in training images by balancing between batch and instance normalizations based on the current task. Though the BIN is designed to learn the trade-off between keeping or reducing original training style variations using simple statistics-based operations, the work in~\cite{nam2018batch} does not provide any study regarding incorrect WB settings. The augmentation and pre-processing methods proposed in our work directly target training and testing images and do not require any change to a DNNs architecture or training regime.

\section{Effects of WB Errors on Pre-trained DNNs} \label{subsec:pre-trained_evaluation}

\begin{figure}
\begin{center}
\includegraphics[width=0.97\linewidth]{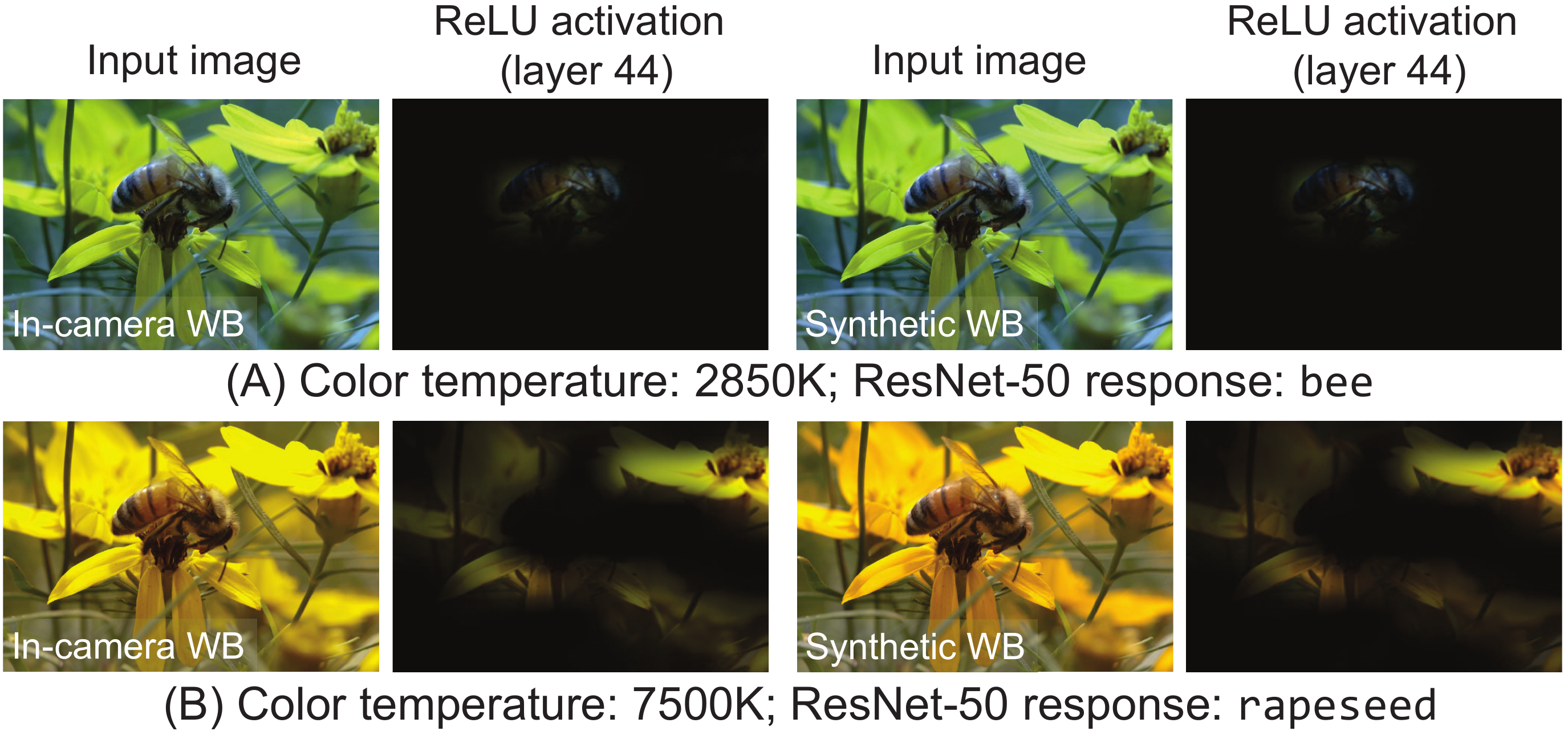}
\vspace{-3mm}
\end{center}
\caption{Image rendered with two different color temperatures (denoted by $t$) using in-camera rendering and our method. (A) Image class is \texttt{bee}. (B) Image class is \texttt{rapeseed}. Classification results were obtained by ResNet-50 \cite{he2016deep}.}
\label{fig:attention}
\end{figure}

We begin by studying the effect of incorrectly white-balanced images on pre-trained DNN models for image classification and semantic segmentation. As a motivation, Fig. \ref{fig:attention} shows two different WB settings applied to the same image.  Fig. \ref{fig:attention} shows that the DNN's attention for the same scene is considerably altered by changing the WB setting.

\begin{figure}[b]
\begin{center}
\includegraphics[width=\linewidth]{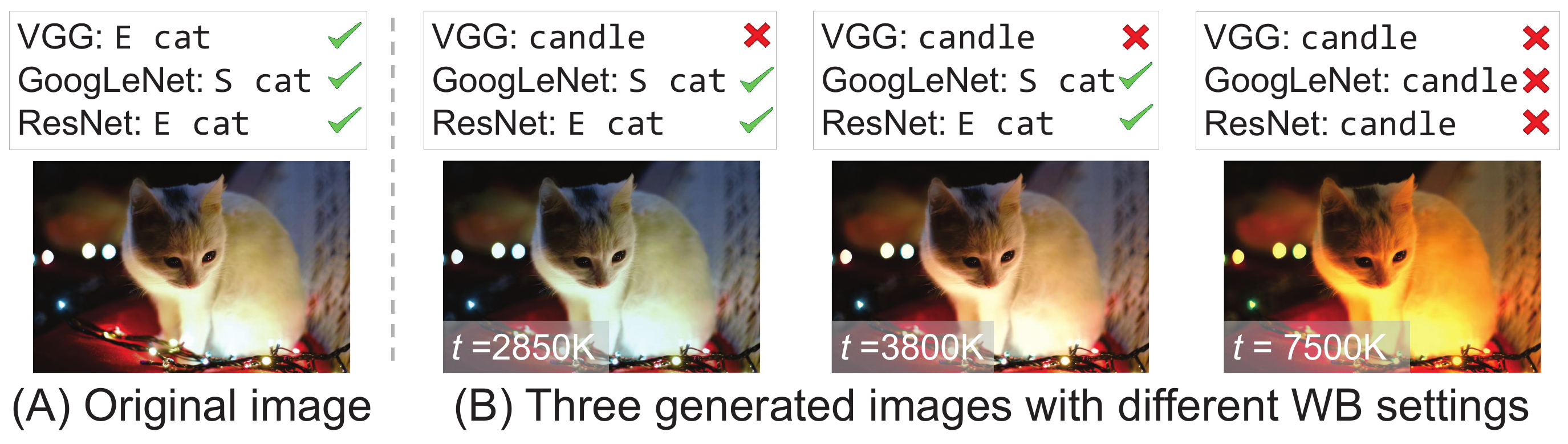}
\vspace{-7mm}
\end{center}
\caption{Pre-trained models are negatively impacted by incorrect WB settings. (A) Original image. (B) Generated images with different WB color temperatures (denoted by $t$). Classification results of: VGG-16 \cite{simonyan2014very}, GoogLeNet \cite{szegedy2015going}, and ResNet-50 \cite{he2016deep} are written on top of each image. The terms \texttt{E} and \texttt{S} stand for \texttt{Egyptian} and \texttt{Siamese}, respectively.}
\label{fig:effect_synthWB_pre_trained}
\end{figure}

For quantitative evaluations, we adopted several DNN models trained for the ImageNet Large Scale Visual Recognition Challenge (ILSVRC) 2012 \cite{deng2009imagenet} and the ADE20K Scene Parsing Challenge 2016 \cite{zhou2017scene}. Generating an entirely new labeled testing set composed of images with incorrect WB is an enormous task---ImageNet classification includes 1,000 classes and pixel-accurate semantic annotation requires $\sim$60 minutes per image \cite{richter2016playing}.   In lieu of a new testing set, we applied our method which emulates WB errors to the validation images of each dataset.  Our method will be detailed shortly in Sec.~\ref{sec:method}.

\paragraph{Classification}~We apply our method to ImageNet's validation set to generate images with five different color temperatures and two different photo-finishing styles for a total of ten WB variations for each validation image; 899 grayscale images were excluded from this process. In total, we generated 491,010 images. We examined the following six well-known DNN models, trained on the original ImageNet training images: AlexNet \cite{krizhevsky2012imagenet}, VGG-16 \& VGG-19 \cite{simonyan2014very}, GoogLeNet \cite{szegedy2015going}, and ResNet-50 \& ResNet-101 \cite{he2016deep}. Table \ref{Table:resultsOfPretrained-classification} shows the accuracy drop for each model when tested on our generated validation set (i.e., with different WB and photo-finishing settings) compared to the original validation set.  In most cases, there is a drop of $\sim$10\% in accuracy. Fig. \ref{fig:effect_synthWB_pre_trained} shows an example of the impact of incorrect WB.

\paragraph{Semantic Segmentation} We used the ADE20K validation set for 2,000 images, and generated ten images with different WB/photo-finishing settings for each image. At the end, we generated a total of 20,000 new images. We tested the following two DNN models trained on the original ADE20K training set: DilatedNet \cite{chen2018deeplab, yu2015multi} and RefineNet \cite{lin2017refinenet}. Table \ref{Table:resultsOfPretrained-segmentation} shows the effect of improperly white-balanced images on the intersection-over-union (IoU) and pixel-wise accuracy (pxl-acc) obtained by the same models on the original validation set.
While DNNs for segmentation fare better than the results for  classification, we still incur a drop of over 2\% in performance.

\begin{table}
\centering
\caption{Adverse performance on ImageNet\cite{deng2009imagenet} due to the inclusion of incorrect WB versions of its validation images. The models were trained on the original ImageNet training set. The reported numbers denote the changes in the top-1 accuracy achieved by each model.}
\vspace{2mm}
\scalebox{0.7}
{
\begin{tabular}{|c|c|}
\hline
\textbf{Model}    & \textbf{Effect on top-1 accuracy} \\\hline
AlexNet \cite{krizhevsky2012imagenet} &  -0.112 \\\hline
VGG-16 \cite{simonyan2014very} &  -0.104 \\\hline
VGG-19 \cite{simonyan2014very}& -0.102  \\\hline
GoogLeNet \cite{szegedy2015going}& -0.107  \\\hline
ResNet-50 \cite{he2016deep} & -0.111 \\\hline
ResNet-101 \cite{he2016deep} & -0.109 \\ \hline
\end{tabular}\label{Table:resultsOfPretrained-classification}
}
\end{table}

\begin{table}
\centering
\caption{Adverse performance on ADE20K~\cite{zhou2017scene} due to the inclusion of incorrect WB versions of its validation images. The models were trained on ADE20K's original training set. The reported numbers denote the changes in intersection-over-union (IoU) and pixel-wise accuracy (pxl-acc) achieved by each model on the original validation.}
\vspace{2mm}
\scalebox{0.7}{
\begin{tabular}{|c|c|c|}
\hline
\textbf{Model} & \textbf{Effect on IoU} & \textbf{Effect on pxl-acc} \\ \hline
 DilatedNet \cite{chen2018deeplab, yu2015multi}  & -0.023                 & -0.024                                                                      \\ \hline
RefineNet \cite{lin2017refinenet}      & -0.031                 & -0.026                                                                      \\ \hline
\end{tabular}
\label{Table:resultsOfPretrained-segmentation}
}
\end{table}

\section{Proposed Method to Emulate WB Errors} \label{sec:method}

\begin{figure*}
\begin{center}
\includegraphics[width=\linewidth]{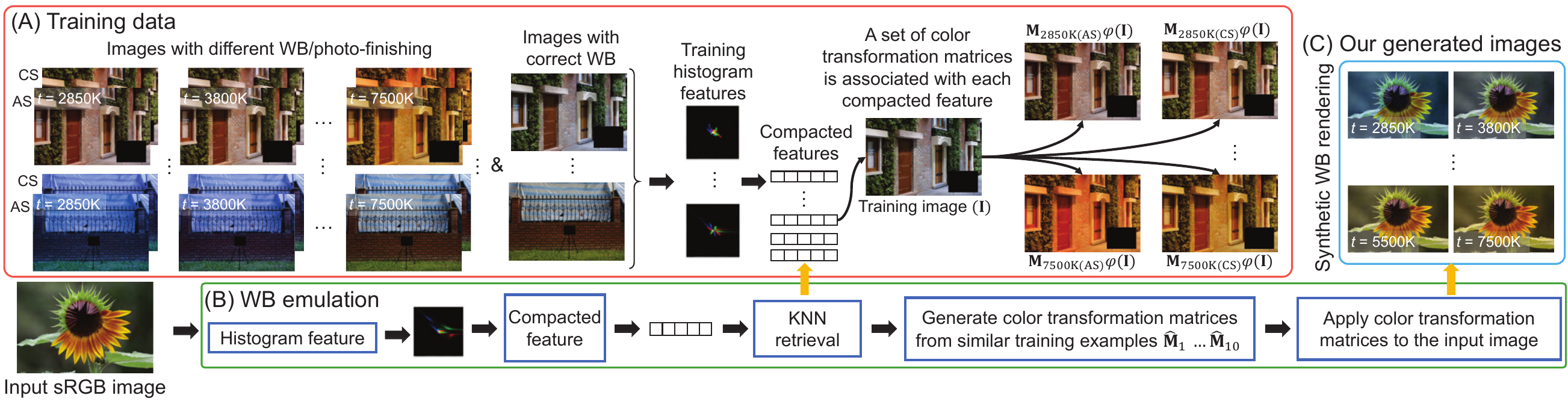}
\vspace{-8mm}
\end{center}
   \caption{Our WB emulation framework. (A) A dataset of 1,797 correctly white-balanced sRGB images \cite{afifi2019color}; each image has ten corresponding sRGB images rendered with five different color temperatures and two photo-finishing styles, Camera Standard (CS) and Adobe Standard (AS). For each white-balanced image, we generate its compact histogram feature and ten color transformation matrices to the corresponding ten images. (B) Our WB emulation pipeline (detailed in Sec. \ref{sec:method}). (C) The augmented images for the input image that represent different color temperatures (denoted by $t$) and photo-finishing styles.}\vspace{-2mm}
\label{fig:overview}
\end{figure*}

Given an sRGB image, denoted as $\mathbf{I}_{t_\textrm{corr}}$, that is assumed to be white-balanced with the correct color temperature, our goal is to modify $\mathbf{I}_{t_\textrm{corr}}$'s colors to mimic its appearance as if it were rendered by a camera with different (incorrect) color temperatures, $t$, under different photo-finishing styles.  Since we do not have access to $\mathbf{I}_{t_\textrm{corr}}$'s original raw-RGB image, we cannot re-render the image from raw-RGB to sRGB using a standard camera pipeline.  Instead, we have adopted a data-driven method that mimics this manipulation directly in the sRGB color space. Our framework draws heavily from the WB-sRGB data-driven framework \cite{afifi2019color}, which was proposed to correct improperly white-balanced sRGB images. Our framework, however, ``emulates'' WB errors on the rendered sRGB images. Fig.~\ref{fig:overview} provides an overview of our method.

\subsection{Dataset} \label{subsec:dataset}

Our method relies on a large dataset of sRGB images generated by \cite{afifi2019color}. This dataset contains images rendered
with different WB settings and photo-finishing styles.
There is a ground truth sRGB image (i.e., rendered with the ``correct'' color temperature) associated with each training image.
The training sRGB images were rendered using five different color temperatures: 2850 Kelvin (K), 3800K, 5500K, 6500K, and 7500K. In addition, each image was rendered using different camera photo-finishing styles.
In our WB emulation framework, we used 17,970 images from this dataset (1,797 correct sRGB images each with ten corresponding images rendered with five different color temperatures and two different photo-finishing styles, Camera Standard and Adobe Standard).

\subsection{Color Mapping} \label{subsec:WBaug}

Next, we compute a mapping between the correct white-balanced sRGB image to each of its ten corresponding images.  We follow the same procedure of the WB-sRGB method \cite{afifi2019color} and use a kernel function, $\varphi$, to project RGB colors into a high-dimensional space. Then, we perform polynomial data fitting on these projected values. Specifically, we used
$\varphi$:$[\textrm{R}$, $\textrm{G}$, $\textrm{B}]^T \rightarrow [\textrm{R}$, $\textrm{G}$, $\textrm{B}$, $\textrm{RG}$, $\textrm{RB}$, $\textrm{GB}$, $\textrm{R}^2$, $\textrm{G}^2$, $\textrm{B}^2]^T$ \cite{finlayson2015color}. The data fitting can be represented by a color transformation matrix $\mathbf{M}_{t_\textrm{corr}\rightarrow t}$ computed by the following minimization equation:
\begin{equation}
\label{eq1_M}
\underset{\mathbf{M}_{t_\textrm{corr}\rightarrow t}}{\argmin} \left\|\mathbf{M}_{t_\textrm{corr}\rightarrow t}\textrm{ }\varphi\left(\mathbf{I}_{t_\textrm{corr}}\right)    - \mathbf{I}_{t}\right\|_{\textrm{F}},
\end{equation}
where $\mathbf{I}_{t_\textrm{corr}}$ and $\mathbf{I}_{t}$ are $3\!\times\!n$ color matrices of the white-balanced image rendered with the correct color temperature $t_\textrm{corr}$ and color values of the same image rendered with the target different color temperature $t$, respectively, $n$ is the total number of pixels in each image, $\left\|.\right\|_{\textrm{F}}$ is the Frobenius norm, and $\mathbf{M}_{t_\textrm{corr}\rightarrow t}$ is represented as a nonlinear $3\!\times\!9$ full matrix.

We compute a color transformation matrix between each pair of correctly white-balanced image and its corresponding target image rendered with a specific color temperature and photo-finishing.
In the end, we have \textit{ten} matrices associated with each image in our training data.

\subsection{Color Feature} As shown in Fig.~\ref{fig:overview}, when augmenting an input sRGB image to have different WB settings, we search our dataset for similar sRGB images to the input image.  This search is not based on scene content, but on the color distribution of the image.  As a result, we represent each image in the training set with the RGB-$uv$ projected color histogram feature used in \cite{afifi2019color}. Each histogram feature is represented as an $m\!\times\!m\!\times\!3$ tensor. To further reduce the size of the histogram feature, we apply principal component analysis (PCA) to the three-layer histogram feature. This transformation maps the zero-centered vectorized histogram to a new lower-dimensional space.  Our implementation used a 55-dimensional PCA vector. Our final training data therefore consists of the compacted feature vector of each training white-balanced image, the associated color transformation matrices, and the PCA coefficient matrix and bias vector.

\subsection{KNN Retrieval} Given a new input image $\mathbf{I}_\textrm{in}$, we extract its compacted color feature $\mathbf{v}$, and then search for training examples with color distributions similar to the input image's color distribution. The $\textrm{L}_2$ distance is adopted as a similarity metric between $\mathbf{v}$ and the training compacted color features. Afterwards, we retrieve the color transformation matrices associated with the nearest $k$ training images. The retrieved set of matrices is represented by  $\mathbf{M}_{\textrm{s}} = \{\mathbf{M}^{(j)}_{\textrm{s}}\}_{j=1}^{j=k}$, where $\mathbf{M}^{(j)}_{\textrm{s}}$ represents the color transformation matrix that maps the $j^{\textrm{th}}$ white-balanced training image colors to their corresponding image colors rendered with color temperature $t$.

\subsection{Transformation Matrix} After computing the distance vector $\mathbf{d}$ between $\mathbf{v}$ and the nearest training features, we compute a weighting vector $\boldsymbol{\alpha}$ as follows \cite{afifi2019color}:

\begin{equation}
\label{eq:weighting}
\boldsymbol{\alpha}_j = \frac{\exp\left(-\mathbf{d}_j^2/2\sigma^2\right)}{\sum_{k^{'}=1}^{k}\exp\left(-\mathbf{d}_{k^{'}}^2/2\sigma^2\right)}, \text{ } j\in[1,...,k],
\end{equation}

\noindent where $\sigma$ is the radial basis function parameter. We used $\sigma=0.25$ in our experiments. We construct the final color transformation matrix $\hat{\mathbf{M}}_{t_\textrm{corr}\rightarrow t}$ as a linear weighted combination of the retrieved color transformation matrices $\mathbf{M}_{\textrm{s}}$.  This process is performed as follows \cite{afifi2019color}:
\begin{equation} \label{eq:final_matrix}
\hat{\mathbf{M}}_{t_\textrm{corr}\rightarrow t} = \sum_{j=1}^{k}{\boldsymbol{\alpha}_j \mathbf{M}^{(j)}_{\textrm{s}}}.
\end{equation}
Lastly, the ``re-rendered'' image $\hat{\mathbf{I}}_{t}$ with color temperature $t$ is computed as:
\begin{equation}
\label{eq:manipulation}
\hat{\mathbf{I}}_{t} = \hat{\mathbf{M}}_{t_\textrm{corr}\rightarrow t}\textrm{ } \varphi\left(\mathbf{I}_\textrm{in}\right).
\end{equation}

\section{Experiments}

\begin{figure}[b]
\begin{center}
\includegraphics[width=0.98\linewidth]{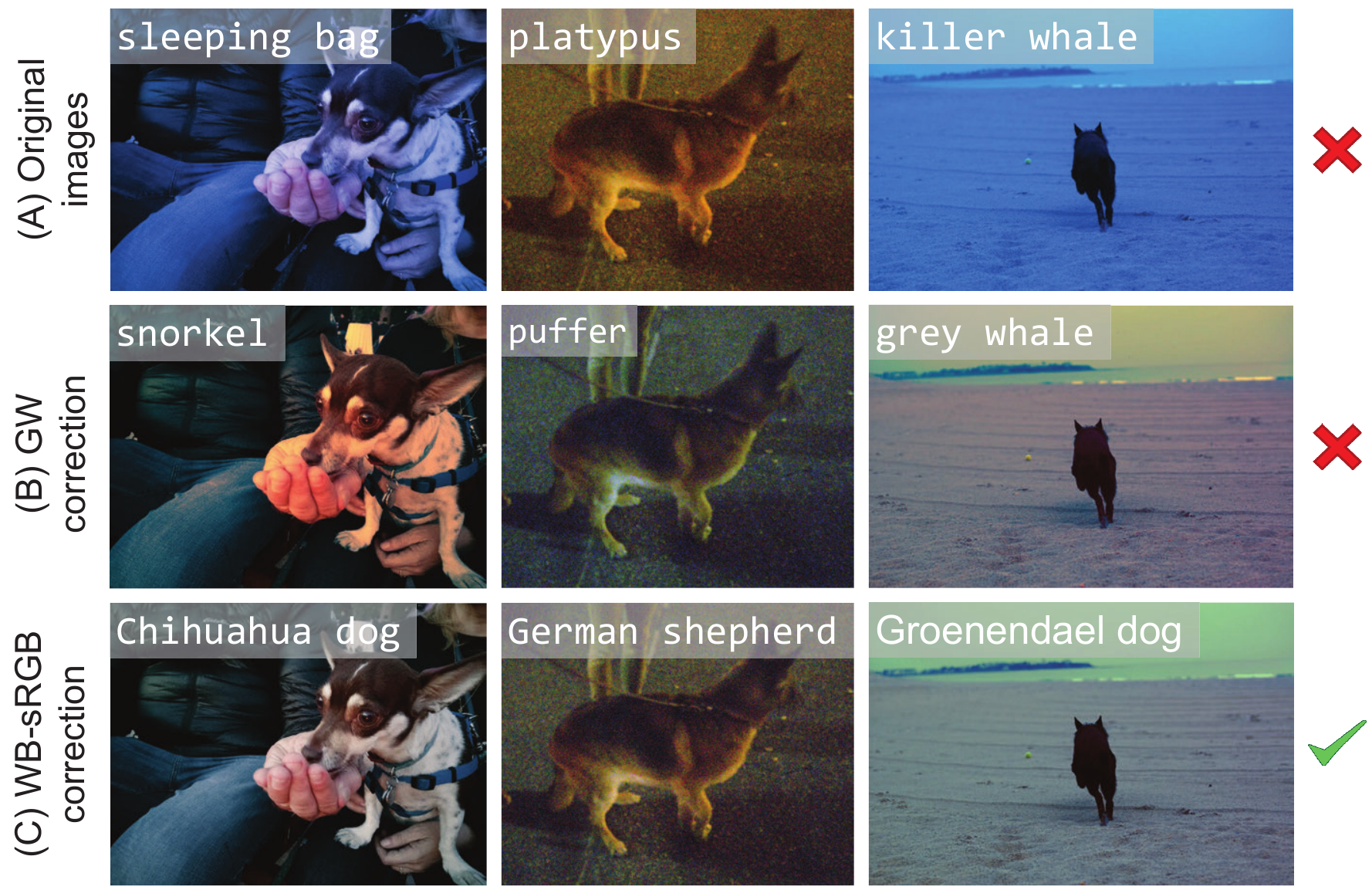}
\end{center}
\vspace{-3mm}
   \caption{(A) Images with different categories of ``dogs'' rendered with incorrect WB settings. (B) Corrected images using gray-world (GW) \cite{GW}. (C) Corrected images using the WB-sRGB method \cite{afifi2019color}. Predicted class by AlexNet is written on top of each image. Images in (A) and (B) are misclassified.}
\label{fig-1}
\end{figure}

\paragraph{Robustness Strategies}~Our goal is to improve the performance of DNN methods in the face of test images that may have strong global color casts due to computational color constancy errors. Based on the WB-sRGB framework \cite{afifi2019color} and the modified framework discussed in Sec. \ref{sec:method}, we examine three strategies to improve the robustness of the DNN models.
\\\\
\noindent \textbf{(1)}~The first strategy is to apply a WB correction to each testing image in order to remove any unexpected color casts during the inference time.  Note that this approach implicitly assumes that the training images are correctly WB. In our experiments, we used the WB-sRGB method \cite{afifi2019color} to correct the test images, because it currently achieves the state-of-the-art on white balancing sRGB rendered images. We examined adapting the simple diagonal-based correction -- which is applied by traditional WB methods that are intended to be applied on raw-RGB images (e.g., gray-world \cite{GW}) -- but found that they give inadequate results when applied on sRGB images, as also demonstrated in \cite{afifi2019color}. In fact, applying diagonal-based correction directly on the training image is similar to multiplicative color jittering. This is why we need to use a nonlinear color manipulation (e.g., polynomial correction estimated by \cite{afifi2019color}) for more accurate WB correction for sRGB images. An example of the difference is shown in Fig. \ref{fig-1}.

It is worth mentioning that the training data used by the WB-sRGB method has five fixed color temperatures (2850K, 3800K, 5500K, 6500K, 7500K), all with color correction matrices mapping to their corresponding correct WB.  In most cases, one of these five fixed color temperatures will be visually similar to the correct WB.  Thus, if the WB-sRGB method is applied to an input image that is already correctly white-balanced, the computed transformation will act as an identity.
\\\\
\noindent \textbf{(2)}~The second strategy considers the case that some of the training images may include some incorrectly white-balanced images. We, therefore, also apply the WB correction step to all the training images as well as testing images. This again uses the WB-sRGB method \cite{afifi2019color} on both testing and training images.
\\\\
\noindent \textbf{(3)}~The final strategy is to augment the training dataset based on our method described in Sec.~\ref{sec:method}. Like other augmentation approaches, there is no pre-processing correction required. The assumption behind this augmentation process is that the robustness of DNN models can be improved by training on augmented images that serve as exemplars for color constancy errors.

\paragraph{Testing Data Categories}~Testing images are grouped into two categories. In Category 1 (Cat-1), we expand the original testing images in the CIFAR-10, CIFAR-100, and ADE20K datasets by applying our method to emulate camera WB errors (described in Sec.~\ref{sec:method}).  Each test image now has ten (10) variations that share the same ground truth labels.  We acknowledge this is less than optimal, given that the same method to modify the testing image is used to augment the training images.  However, we are confident in the proposed method's ability to emulate WB errors that we feel Cat-1 images represents real-world examples.  With that said,  we do not apply strategies 1 and 2 to Cat-1, as the WB-sRGB method is based on a similar framework used to generate the testing images.   For the sake of completeness, we also include Category  2 (Cat-2), which consists of new datasets generated directly from raw-RGB images.  Specifically, raw-RGB images are rendered using the full in-camera pipeline to sRGB images with in-camera color constancy errors. As a result, Cat-2's testing images exhibit accurate color constancy errors but contain fewer testing images for which we have provided the ground truth labels.

\subsection{Experimental Setup}

We compare the three above strategies with two existing and widely adopted color augmentation processes:  RGB color jittering and HSV jittering.

\paragraph{Our Method} The nearest neighbor searching was applied using $k=25$.
The proposed WB augmentation model runs in 7.3 sec (CPU) and 1.0 sec (GPU) to generate \textit{ten} 12-mega-pixel images.
The reported runtime was computed using Intel$^\circledR$ Xeon$^\circledR$ E5-1607 @ 3.10 GHz CPU and $\textrm{NVIDIA\texttrademark}$ Titan X GPU.

\paragraph{Existing Color Augmentation}
To the best of our knowledge, there is no standardized approach for existing color augmentation methods. Accordingly, we tested different settings and selected the settings that produce the best results.

For RGB color jittering, we generated ten images with new colors by applying a random shift $x \sim \mathcal{N}(\mu_x,\sigma^2)$ to each color channel of the image. For HSV jittering, we generated ten images with new colors by applying a random shift $x$ to the hue channel and multiplying each of the saturation and value channels by a random scaling factor $s \sim \mathcal{N}(\mu_s,\sigma^2)$. We found that $\mu_x = -0.3$, $\mu_s = 0.7$, and $\sigma = 0.6$ give us the best compromise between having color diversity with low color artifacts during the augmentation process.

\subsection{Network Training} \label{subsec:training}

For image classification, training new models on ImageNet dataset requires unaffordable efforts---for instance, ILSVRC 2012 consists of $\sim$1 million images and would be $\sim$10 million images after applying any of the color augmentation techniques. For that reason, we perform experiments on CIFAR-10 and CIFAR-100 datasets \cite{krizhevsky2009learning} due to a more manageable number of images in each dataset.

We trained SmallNet \cite{perez2017effectiveness} from scratch on CIFAR-10. We also fine-tuned AlexNet \cite{krizhevsky2012imagenet} to recognize the new classes in CIFAR-10 and CIFAR-100 datasets. For semantic segmentation, we fine-tuned SegNet \cite{badrinarayanan2017segnet} on the training set of the ADE20K dataset \cite{zhou2017scene}.

We train each model on: (i) the original training images, (ii) the WB-sRGB method \cite{afifi2019color} applied to the original training images, and (iii) original training images with the additional images produced by color augmentation methods.
For color augmentation, we examined RGB color jittering, HSV jittering, and our WB augmentation. Thus, we trained five models for each CNN architecture, each of which was trained on one of the mentioned training settings.

For fair comparisons, we trained each model for the same number of iterations. Specifically, the training was for $\sim$29,000 and $\sim$550,000 iterations for image classification and semantic segmentation tasks, respectively. We adjusted the number of epochs to make sure that each model was trained on the same number of mini-batches for fair comparison between training on augmented and original sets.
Note that by using a fixed number of iterations to train models with both original training data and augmented data, we did not fully exploit the full potential of the additional training images when we trained models using additional augmented data.

The training was performed using $\textrm{NVIDIA\texttrademark}$ Titan X GPU. The details of training parameters are given in supplemental materials.

\subsection{Results on Cat-1}

Cat-1 tests each model using test images that have been generated by our method described in Sec.~\ref{sec:method}.

\paragraph{Classification} We used the CIFAR-10 testing set (10,000 images) to test SmallNet and AlexNet models trained on the training set of the same dataset. We also used the CIFAR-100 testing set (10,000 images) to evaluate the AlexNet model trained on CIFAR-100. After applying our WB emulation to the testing sets, we have 100,000 images for each testing set of CIFAR-10 and CIFAR-100. The top-1 accuracies obtained by each trained model are shown in Table \ref{Table:resultsOfSmallNetAlexNet}. The best results on our expanded testing images, which include strong color casts, were obtained using models trained on our proposed WB augmented data.

Interestingly, the experiments show that applying WB correction \cite{afifi2019color} on the training data, in most cases, improves the accuracy using both the original and expanded test sets. DNNs that were trained on WB augmented training images achieve the best improvement on the original testing images compared to using other color augmenters.

\begin{table}[]
\centering
\caption{[Cat-1] Results of SmallNet \cite{perez2017effectiveness} and AlexNet \cite{krizhevsky2012imagenet} on CIFAR dataset \cite{krizhevsky2009learning}. The shown accuracies obtained by models trained on: original training, ``white-balanced'', and color augmented sets. The testing was performed using: original testing set and testing set with different synthetic WB settings (denoted as diff. WB).
The results of the baseline models (i.e., trained on the original training set) are highlighted in green, while the best result for each testing set is shown bold. We highlight best results obtained by color augmentation techniques in yellow. Effects on baseline model results are shown in parentheses. \label{Table:resultsOfSmallNetAlexNet}}
\vspace{3mm}
\scalebox{0.7}
{
\begin{tabular}{c|c|c|}
\cline{2-3}
\multicolumn{1}{c|}{\textbf{Cat-1}} & \multicolumn{2}{c|}{\textbf{SmallNet \cite{perez2017effectiveness} on CIFAR-10 \cite{krizhevsky2009learning}}} \\ \hline
\multicolumn{1}{|c|}{\textbf{Training set}} & Original & Diff. WB \\ \hline
\multicolumn{1}{|c|}{Original training set} & \cellcolor[HTML]{9AFF99}0.799 &\cellcolor[HTML]{9AFF99} 0.655 \\ \hline
\multicolumn{1}{|c|}{``White-balanced'' set} & 0.801 (+0.002) & 0.683 (+0.028) \\ \hline
\multicolumn{1}{|c|}{HSV augmented set} & 0.801 (+0.002) & 0.747 (+0.092) \\ \hline
\multicolumn{1}{|c|}{RGB augmented set} & 0.780 (-0.019) & 0.765 (+0.11) \\ \hline
\multicolumn{1}{|c|}{WB augmented set (ours)} & \cellcolor[HTML]{FFFC9E}\textbf{0.809 (+0.010)} &  \cellcolor[HTML]{FFFC9E}\textbf{0.786 (+0.131)} \\ \hline
\cline{2-3}
\multicolumn{1}{c|}{\textbf{Cat-1}} & \multicolumn{2}{c|} {\textbf{AlexNet \cite{krizhevsky2012imagenet} on CIFAR-10 \cite{krizhevsky2009learning}}} \\ \hline
\multicolumn{1}{|c|}{Original training set} & \cellcolor[HTML]{9AFF99}\textbf{0.933} & \cellcolor[HTML]{9AFF99} 0.797 \\ \hline
\multicolumn{1}{|c|}{``White-balanced'' set} & 0.932 (-0.001) & 0.811 (+0.014) \\ \hline
\multicolumn{1}{|c|}{HSV augmented set} & 0.923 (-0.010) & 0.864 (+0.067) \\ \hline
\multicolumn{1}{|c|}{RGB augmented set} & 0.922 (-0.011) & 0.872 (+0.075) \\ \hline
\multicolumn{1}{|c|}{WB augmented set (ours)} & \cellcolor[HTML]{FFFC9E}0.926 (-0.007) & \cellcolor[HTML]{FFFC9E}\textbf{0.889 (+0.092)} \\ \hline
\cline{2-3}
\multicolumn{1}{c|}{\textbf{Cat-1}} & \multicolumn{2}{c|} {\textbf{AlexNet \cite{krizhevsky2012imagenet} on CIFAR-100 \cite{krizhevsky2009learning}}} \\ \hline
\multicolumn{1}{|c|}{Original training set} & \cellcolor[HTML]{9AFF99}\textbf{0.768} & \cellcolor[HTML]{9AFF99}0.526 \\ \hline
\multicolumn{1}{|c|}{``White-balanced'' set} & 0.757 (-0.011) & 0.543 (+0.017) \\ \hline
\multicolumn{1}{|c|}{HSV augmented set} & 0.722 (-0.044) & 0.613 (+0.087) \\ \hline
\multicolumn{1}{|c|}{RGB augmented set} & 0.723 (-0.045) & 0.645 (+0.119) \\ \hline
\multicolumn{1}{|c|}{WB augmented set (ours)} & \cellcolor[HTML]{FFFC9E}0.735 (-0.033) & \cellcolor[HTML]{FFFC9E}\textbf{0.670 (+0.144)} \\ \hline
\end{tabular}}
\end{table}

\paragraph{Semantic Segmentation} We used the ADE20K validation set using the same setup explained in Sec. \ref{subsec:pre-trained_evaluation}. Table \ref{Table:resultsOfSegNEt} shows the obtained pxl-acc and IoU of the trained SegNet models. The best results were obtained with our WB augmentation; Fig. \ref{fig:segNet} shows qualitative examples. Additional examples are also given in supplemental materials.

\begin{table}[]
\centering
\caption{[Cat-1] Results of SegNet \cite{badrinarayanan2017segnet} on the ADE20K validation set \cite{zhou2017scene}. The shown intersection-over-union (IoU) and pixel-wise accuracy (pxl-acc) were achieved by models trained using: original training, ``white-balanced'', and color augmented sets. The testing was performed using: original testing set and testing set with different synthetic WB settings (denoted as diff. WB). Effects on results of SegNet trained on the original training set are shown in parentheses.  Highlight marks are as described in Table \ref{Table:resultsOfSmallNetAlexNet}.
\label{Table:resultsOfSegNEt}}
\scalebox{0.7}
{
\begin{tabular}{c|c|c|}
\cline{2-3}
 & \multicolumn{2}{c|}{\textbf{IoU}} \\ \cline{2-3}
\multicolumn{1}{c|}{\textbf{Cat-1}} & Original & Diff. WB \\ \hline
\multicolumn{1}{|c|}{Original training set} & \cellcolor[HTML]{9AFF99}0.208 & \cellcolor[HTML]{9AFF99} 0.180  \\ \hline
\multicolumn{1}{|c|}{``White-balanced'' set} & \textbf{0.210 (+0.002)} & 0.197 (+0.017) \\ \hline
\multicolumn{1}{|c|}{HSV augmented set} & 0.192 (-0.016) & 0.185 (+0.005) \\ \hline
\multicolumn{1}{|c|}{RGB augmented set} & 0.195 (-0.013) & 0.190 (+0.010) \\ \hline
\multicolumn{1}{|c|}{WB augmented set (ours)} & \cellcolor[HTML]{FFFC9E}0.202 (-0.006) & \cellcolor[HTML]{FFFC9E}\textbf{0.199 (+0.019)} \\ \hline
\multicolumn{1}{c|}{\textbf{Cat-1}} & \multicolumn{2}{c|}{\textbf{pxl-acc}} \\ \hline
\multicolumn{1}{|c|}{Original training set} & \cellcolor[HTML]{9AFF99}0.603 & \cellcolor[HTML]{9AFF99} 0.557 \\ \hline
\multicolumn{1}{|c|}{``White-balanced'' set} & \textbf{0.605 (+0.002)} & 0.579 (+0.022) \\ \hline
\multicolumn{1}{|c|}{HSV augmented set} & 0.583 (-0.020) & 0.536 (-0.021) \\ \hline
\multicolumn{1}{|c|}{RGB augmented set} & 0.544 (-0.059) & 0.534 (-0.023) \\ \hline
\multicolumn{1}{|c|}{WB augmented set (ours)} & \cellcolor[HTML]{FFFC9E}0.597 (-0.006) & \cellcolor[HTML]{FFFC9E}\textbf{0.581 (+0.024)} \\ \hline
\end{tabular}}

\end{table}

\begin{figure*}
\begin{center}
\includegraphics[width=0.97\linewidth]{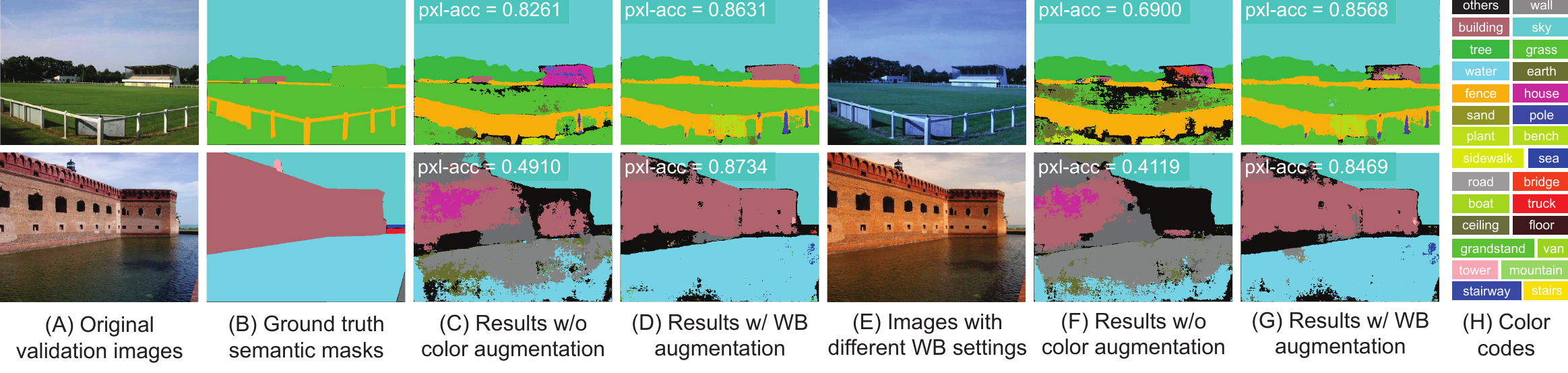}
\vspace{-3.5mm}
\end{center}
   \caption{Results of SegNet \cite{badrinarayanan2017segnet} on the ADE20K validation set \cite{zhou2017scene}. (A) Original validation image. (B) Ground truth semantic mask. (C) \& (D) Results of trained model wo/w color augmentation using image in (A), respectively. (E) Image with a different WB. (F) \& (G) Results w/o and with color augmentation using image in (E), respectively. (H) Color codes. The term `pxl-acc' refers to pixel-wise accuracy.}\vspace{-2.5mm}
\label{fig:segNet}
\end{figure*}

\begin{table}[]
\centering
\caption{[Cat-2] Results of SmallNet \cite{perez2017effectiveness} and AlexNet \cite{krizhevsky2012imagenet}. The shown accuracies were obtained using trained models on the original training, ``white-balanced'', and color augmented sets. Effects on results of models trained on the original training set are shown in parentheses.
Highlight marks are as described in Table \ref{Table:resultsOfSmallNetAlexNet}.
\label{Table:RealDataResults}}
\vspace{2mm}
\scalebox{0.62}{

\begin{tabular}{c|c|c|c|}
\cline{2-4}
\textbf{Cat-2}  & \multicolumn{3}{c|}{\textbf{SmallNet}} \\ \hline
\multicolumn{1}{|c|}{\textbf{Training Set}} & In-cam AWB & In-cam Diff. WB & WB pre-processing \\ \hline
\multicolumn{1}{|c|}{Original training set} & \cellcolor[HTML]{9AFF99}0.467 & \cellcolor[HTML]{9AFF99}0.404 & \cellcolor[HTML]{9AFF99}0.461 \\ \hline
\multicolumn{1}{|c|}{``White-balanced'' set} & \textbf{0.496 (+0.029)} & 0.471 (+0.067) & \textbf{0.492 (+0.031)} \\ \hline
\multicolumn{1}{|c|}{HSV augmented set} & 0.477 (+0.001) & 0.462 (+0.058) & 0.481 (+0.02) \\ \hline
\multicolumn{1}{|c|}{RGB augmented set} & 0.474 (+0.007) & 0.475 (+0.071) & 0.470 (+0.009) \\ \hline
\multicolumn{1}{|c|}{WB augmented set (ours)} & \cellcolor[HTML]{FFFC9E}0.494 (+0.027) & \cellcolor[HTML]{FFFC9E}\textbf{0.496 (+0.092)} & \cellcolor[HTML]{FFFC9E}0.484 (+0.023) \\ \hline
\multicolumn{1}{c|}{\textbf{Cat-2}} & \multicolumn{3}{c|}{\textbf{AlexNet}} \\ \hline
\multicolumn{1}{|c|}{Original training set} & \cellcolor[HTML]{9AFF99}0.792 & \cellcolor[HTML]{9AFF99}0.734 & \cellcolor[HTML]{9AFF99}0.772 \\ \hline
\multicolumn{1}{|c|}{``White-balanced'' set} & 0.784 (-0.008) & 0.757 (+0.023) & 0.784 (+0.012) \\ \hline
\multicolumn{1}{|c|}{HSV augmented set} & 0.790 (+0.002) & 0.771 (+0.037) & 0.779 (+0.007) \\ \hline
\multicolumn{1}{|c|}{RGB augmented set} & 0.791 (-0.001) & 0.779 (+0.045) & 0.783 (+0.011) \\ \hline
\multicolumn{1}{|c|}{WB augmented set (ours)} & \cellcolor[HTML]{FFFC9E}\textbf{0.799 (+0.007)} & \cellcolor[HTML]{FFFC9E}\textbf{0.788 (+0.054)} & \cellcolor[HTML]{FFFC9E}\textbf{0.787 (+0.015)} \\ \hline
\end{tabular}
}\vspace{-1.5mm}
\end{table}

\subsection{Results on Cat-2} \label{subsec:Testing-realdata}

Cat-2 data requires us to generate and label our own testing image dataset using raw-RGB images.   To this end, we collected 518 raw-RGB images containing CIFAR-10 object classes from the following datasets:  HDR+ Burst Photography dataset \cite{hasinoff2016burst}, MIT-Adobe FiveK dataset \cite{bychkovsky2011learning}, and Raise dataset \cite{dang2015raise}. We rendered all raw-RGB images with different color temperatures and two photo-finishing styles using the  Adobe Camera Raw module. Adobe Camera Raw accurately emulates the ISP onboard a camera and produces results virtually identical to what the in-camera processing would produce\cite{afifi2019color}. Images that contain multiple objects were manually cropped to include only the interesting objects---namely, the CIFAR-10 classes. At the end, we generated 15,098 rendered testing images that reflect real in-camera WB settings. We used the following testing sets in our experiments:
\\\\
\noindent\textbf{(i) In-camera auto WB} contains images rendered with the auto WB (AWB) correction setting in Adobe Camera Raw, which mimics the camera's AWB functionality. AWB does fail from time to time; we manually removed images that had a noticeable color cast. This set of images is intended to be equivalent to testing images on existing image classification datasets.
\\\\
\noindent\textbf{(ii) In-camera WB settings} contains images rendered with the different color temperatures and photo-finishing styles.  This set represents testing images that contain WB color cast errors.
\\\\
\noindent\textbf{(iii) WB pre-processing correction applied to set (ii)} contains images of set (ii) after applying the WB-sRGB correction \cite{afifi2019color}. This set is used to study the potential improvement of applying a pre-processing WB correction in the inference phase.

Table \ref{Table:RealDataResults} shows the top-1 accuracies obtained by SmallNet and AlexNet on the external testing sets. The experiments show the accuracy is reduced by $\sim$6\% when the testing set is images that have been modified with  incorrect WB settings compared with their original accuracies obtained with ``properly'' white-balanced images using the in-camera AWB. We also notice that the best accuracies are obtained by applying either a pre-processing WB on both training/testing images or our WB augmentation in an end-to-end manner. Examples of misclassified images are shown in Fig. \ref{fig:realFalseExamples}. Additional examples are also given in supplemental materials.

\begin{figure}
\begin{center}
\includegraphics[width=0.96\linewidth]{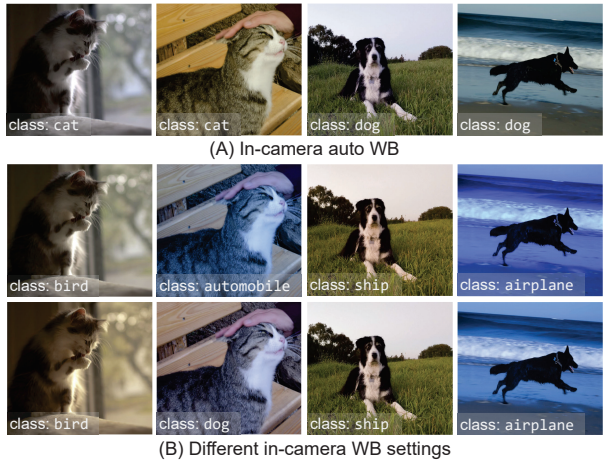}
\end{center}
\vspace{-3.5mm}
   \caption{(A) Correctly classified images rendered with in-camera auto WB. (B) Misclassified images rendered with \textit{in-camera} different WB. Note that all images in (B) are correctly classified by the same model (AlexNet \cite{krizhevsky2012imagenet}) trained on WB augmented data.}\vspace{-1.5mm}
\label{fig:realFalseExamples}
\end{figure}

\section{Conclusion}

This work has examined the impact on computational color constancy errors  on DNNs for image classification and semantic segmentation. A new method to perform augmentation that accurately mimics WB errors was introduced. We show that both pre-processing WB correction and training DNNs with our augmented WB images improve the results for DNNs targeting CIFAR-10, CIFAR-100, and ADE20K datasets.  We believe our WB augmentation method will be useful for other tasks targeted by DNN where image augmentation is sought.

\vspace{-2.5mm}
\small\paragraph{Acknowledgments}  This study was funded in part by the Canada First Research Excellence Fund for the Vision: Science to Applications (VISTA) programme and an NSERC Discovery Grant.  Dr. Brown contributed to this article in his personal capacity as a professor at York University.  The views expressed are his own and do not necessarily represent the views of Samsung Research.

{\small
\bibliographystyle{ieee_fullname}

}

\end{document}